%% file: acl2020.tex
\documentclass[11pt,a4paper]{article}
\usepackage[hyperref]{acl2020}
\usepackage{times}
\usepackage{latexsym}

\usepackage{amsmath}
\usepackage{amssymb}
\usepackage{amsthm}
\usepackage{graphicx}
\usepackage{booktabs}
\graphicspath{ {./images/} }
\theoremstyle{definition}
\newtheorem{definition}{Definition}
\newcommand{\ra}[1]{\renewcommand{\arraystretch}{#1}}
 
\theoremstyle{remark}

\usepackage{microtype}

\aclfinalcopy 
\def\aclpaperid{39} 

\title{RPD: A Distance Function Between Word Embeddings}

\author{
Xuhui Zhou,\textsuperscript{\rm 1}
Zaixiang Zheng,\textsuperscript{\rm 2}
Shujian Huang,\textsuperscript{\rm 2}\\
\textsuperscript{\rm 1}University of Washington\\
\textsuperscript{\rm 2}National Key Laboratory for Novel Software Technology, Nanjing University\\
\texttt{xuhuizh@uw.edu, zhengzx@smail.nju.edu.cn, huangsj@nju.edu.cn} 
}

\date{}

\begin{document}
\maketitle
\begin{abstract}
It is well-understood that different algorithms, training processes, and corpora produce different word embeddings. However, less is known about the relation between different embedding spaces, i.e. how far different sets of embeddings deviate from each other. In this paper, we propose a novel metric called Relative pairwise inner Product Distance~(RPD) to quantify the distance between different sets of word embeddings. This metric has a unified scale for comparing different sets of word embeddings. Based on the properties of RPD, we study the relations of word embeddings of different algorithms systematically, and investigate the influence of different training processes and corpora. The results shed light on the poorly understood word embeddings and justify RPD as a measure of the distance of embedding spaces.
\end{abstract}

\section{Introduction}

Word embeddings are important in Natural language processing~(NLP) which map words into a low-dimensional vector space. Many works have been proposed to generate word embeddings~\cite{Mnih2013LearningWE,Mikolov:13,pennington:14,Levy2014DependencyBasedWE,bojanowski-etal-2017-enriching, devlin-etal-2019-bert}.  

With many different sets of word embeddings produced by different algorithms and corpora, it is interesting to investigate the relationships between these sets of word embeddings. Intrinsically, this would help us better understand word embeddings \citep{levy:15}. Practically, knowing the relationship between different sets of word embeddings helps us build better word meta-embeddings ~\cite{yin-schutze-2016-learning}, reduce biases in word embeddings~\citep{NIPS2016_6228}, pick better hyper-parameters~\cite{Yin:18}, and choose suitable algorithms in different scenarios~\cite{doi:10.1177/0003122419877135}.

To study the relationship between different embedding spaces systematically, we propose RPD as a measure of the distance between different sets of embeddings. We derive statistical properties of RPD including its asymptotic upper bound and normality under the independence condition. We also provide a geometric interpretation of RPD. Furthermore, we show that RPD is strongly correlated with the performance of word embeddings measured by intrinsic metrics, such as comparing semantic
similarity and evaluating analogies.

With the help of RPD, we study the relations among several popular embedding methods, including GloVe~\cite{pennington:14}, SGNS\footnote{Skip-gram with Negative Sampling}~\cite{Mikolov:13}, Singular Value Decomposition~(SVD) factorization of PMI matrix, and SVD factorization of log count~(LC) matrix. Results show that these methods are statistically correlated, which suggests that there is an unified theory behind these methods. 

Additionally, we analyze the influence of training processes, i.e. hyperparameters~(negative sampling), random initialization; and the influence of corpora towards word embeddings. Our findings include the fact that different training corpora result in significantly different GloVe embeddings, and that the main difference between embedding spaces comes from the algorithms although hyperparameters also have certain influence. Those findings not only provide some interesting insights of word embeddings but also fit nicely with our intuition, which further proves RPD as a suitable measure to quantify the relationship between different sets of word embeddings.

\section{Background}
Before introducing RPD, we review the theory behind some static word embedding methods, and discuss some previous works investigating the relationship between embedding spaces.

\subsection{Word Embedding Models}
We consider the following four word embedding models: SGNS, GloVe, SVD$_\text{PMI}$, SVD$_\text{LC}$. SGNS and GloVe are two widely used embedding methods, while SVD$_\text{PMI}$ and SVD$_\text{LC}$ are matrix factorization methods which are intrinsically related to SGNS and GloVe \cite{Levy:14, levy:15, Yin:18}. 

The embedding of all the words forms an embedding matrix $E \in \mathbf{R}^{n \times d}$, where the $d$ here is the dimension of each word vector and $n$ is the size of the vocabulary.

\textbf{SGNS} maximizes a likelihood function for word and context pairs that occur in the dataset and minimizes it for randomly sampled unobserved pairs, i.e. negative samples (NS). We denote the method with $k$ NS as SGNS$_{k}$.

\textbf{GloVe} factorizes the log-count matrix shifted by the entire vocabulary's bias term. The bias here are parameters learned stochastically with an objective weighted according to the frequency of words.

\textbf{SVD$_\text{PMI/LC}$} SVD factorizes a signal matrix $M=UDV^T$, which aims at reducing the dimensions of the 
cooccurrence matrix. The resulting embedding is $E = U_{:,1:d} D^{\frac{1}{2}}_{1:d,1:d}$ , where $d$ is the dimension of word embeddings. We denote the method as SVD$_\text{PMI}$, if the signal is the PMI matrix, and SVD$_\text{LC}$ if the signal is the log count matrix.

Although the scope of this paper focuses 
on standard word embeddings that were learned at the word level, RPD could be adapted to analyze embeddings that were learned from word pieces, for example, fastText \citep{bojanowski-etal-2017-enriching} and contextualized embeddings \citep{peters-etal-2018-deep, devlin-etal-2019-bert}.

\subsection{Relationship Between Embedding Spaces}

\citet{Levy:14} provide a good analogy between SGNS and SVD$_\text{PMI}$. They suggest that SGNS is essentially factorizing the pointwise mutual information~(PMI) matrix. However, their analogy is based on the assumption of no dimension constraint in SGNS, which is not possible in practice. Furthermore, their analogy is not suitable for analyzing methods besides SGNS and PMI models since their theoretical derivation relies on the specific objective of SGNS. 

\citet{Yin:18} provide a way to select the best dimension of word embeddings for specific tasks by exploring the relations of embedding spaces of different dimension. They introduce Pairwise Inner Product (PIP) loss~\cite{Yin:18}, an unitary-invariant metric for measuring word embeddings' distance~\cite{smith:17}. The unitary-invariance of word embeddings states that two embedding vector spaces are equivalent if one can be obtained from another by multiplying a unitary matrix. However, PIP loss is not suitable for comparing numerically across embedding spaces since PIP loss has different energy for different embedding spaces.

\section{Quantifying Distances between Embeddings}
 In this section, we describe the definition of RPD and its properties, which make RPD a suitable and effective method to quantify the distance between embedding spaces. Note that two embedding spaces do not necessarily have the same vocabulary for calculating the RPD.  

\subsection{RPD}
 
For the following discussion, we always use the Frobenius norm as the norm of matrices. 

\theoremstyle{definition}
\begin{definition}{(RPD)}
The RPD between embedding matrices $E_1$ and $E_2$ is defined as follows: 
\[\text{RPD}(E_1, E_2) = \frac{1}{2}\frac{\| \tilde {E_1}\tilde{E_1}^T-\tilde{E_2}\tilde{E_2}^T\|^2}{\|\tilde {E_1}\tilde{E_1}^T\|\|\tilde{E_2}\tilde{E_2}^T\|}.\]
\end{definition}
where $\tilde{E}$ comes from dividing each entry of $E$ by its standard deviation. For convenience, we let $\tilde{E} \equiv E$ for the following discussion.

The numerator of RPD respects the unitary-invariant property of word embeddings, which means that unitary transformation (i.e. rotation) preserves the relative geometry of an embedding space. The denominator is a normalization, which allows us to regard the whole embedding matrix as an integrated part (i.e. RPD does not correlate with the number of words of embedding spaces). This step makes comparisons across methods possible.

\subsection{Statistical Properties of RPD}
We assume the widely used isotropic assumption~\cite{arora:16} that the ensemble of word vectors consists of i.i.d draws generated by $v = s\hat{v}$, where $\hat{v}$ is from the spherical Gaussian distribution, and $s$ is a scalar random variable. In our case, we can assume each entry of embedding comes from a standard normal distribution $E$: $v_{ij} \sim \mathcal{N}(0,1)$. 

Note that the assumption may not always work in practice, especially for other embeddings such as contextualized embeddings. However, under the isotropic conditions, the statistical properties derived are intuitively and empirically plausible. Besides, those properties serve to better interpret the value of RPD alone. Since RPD, in many cases, is used for comparison, we should be comfortable with the assumption. 

\noindent\textbf{Upper bound} We estimate the asymptotic upper bound of RPD. By factorizing the numerator of RPD, we get (1).
\begin{multline}
    \text{RPD}(E_1,E_2)=\frac{1}{2}\frac{\|E_1E_1^T\|^2+\|E_2E_2^T\|^2}{\|E_1E_1^T\|\|E_2E_2^T\|}\\
    -\frac{\langle E_1E_1^T, E_2E_2^T\rangle}{\|E_1E_1^T\|\|E_2E_2^T\|}
\end{multline}
Applying the Cauchy-Schwarz inequality to the last term of (1)\footnote{The inner product of matrix A and B is defined as $\langle\,A,B\rangle=trace(A^TB)$}, we have the following estimation.
\begin{equation}
\begin{split}
  2\text{RPD}(E_1,E_2)  
  &\leq \frac{\|E_1E_1^T\|^2+\|E_2E_2^T\|^2}{\|E_1E_1^T\|\|E_2E_2^T\|}\\
  &= \frac{\|E_1E_1^T\|}{\| E_2E_2^T\|} + \frac{\|E_2E_2^T\|}{\|E_1E_1^T\|} 
\end{split}
\end{equation}
By the law of large numbers, we can prove that $\lim_{n\to\infty}\|EE^T\| = n\sqrt{d}$~(Appendix A).
Then, we can tell from (2) that RPD is bounded by 1 when $n \to \infty$. In practice, the number of words $n$ is large enough to let the maximum of RPD stay around 1, which means RPD is well-defined numerically.\\

\noindent\textbf{Normality} 
For $\text{RPD}(E_1,E_2)$, if $E_1$ is independent of $E_2$, we can prove that RPD distributes normally from both an empirical and a theoretical perspective. Theoretically, by applying the central limit theorem to the numerator and the law of large numbers to the denominator of RPD, we can get the normality of RPD under the condition $n \to \infty$, $\frac{d}{n}=c$, where $c$ remains constant~(Appendix B). Empirically, we can use Monte Carlo simulation to show the normality and estimate the mean and variance of RPD~(Appendix C).
With the help of RPD, we can perform hypothesis test~(z-test) to evaluate the independence of two embedding spaces.

\subsection{Geometric Interpretation of RPD}

From equation (1), we can tell that the first term goes to 1 when $n \to \infty$. So we only need to discuss the second term. 
\[\frac{\langle E_1E_1^T, E_2E_2^T\rangle}{\|E_1E_1^T\|\|E_2E_2^T\|}\]

For the $i^{th}$ row in $EE^T$, we have vector $\hat{v}_i =(v_iv_1^T,v_iv_2^T,... ,v_iv_n^T)$, where $v_i$ is the word $i$'s vector in embedding $E$, $n$ is the number of words. We can interpret $\hat{v}_i$ as another representation of word i projected onto the space spanned by $v_1, v_2, ..., v_n$. So for convenience, we denote $\hat{E} = EE^T$ with its $i^{th}$ row as $\hat{v}_i$.

We can prove that
$\lim_{n\to \infty}\text{RPD}(E_1,E_2) = 1-\frac{1}{n}\sum_{i=1}^n \cos(\theta_i)$. The $\theta_i \in (0, \frac{\pi}{2})$ is the angle between $\hat{v}_i^{(1)}$~($i^{th}$ row vector of $\hat{E_1}$) and $\hat{v}_i^{(2)}$~($i^{th}$ row vector of $\hat{E_2}$)~(Appendix D). Therefore, we can understand the value of RPD from the perspective of cosine similarity between vectors.

\begin{figure}[h]
\centering
\includegraphics[width=0.45\textwidth]{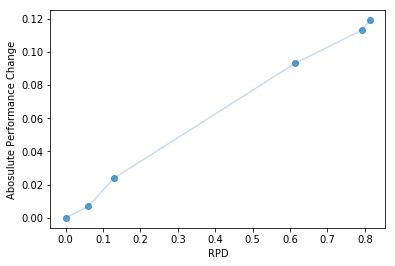}
\caption{The plot shows the difference in performance as a function of RPD 
score. The x-axis for each point represents the RPD between  word embeddings produced by SGNS (with NS 15, 5, 1), GloVe, SVD$_\text{PMI}$, SVD$_\text{LC}$ and word embeddings produced by SGNS$_{25}$. The y-axis for each point represents the sum of absolute variation in the performance (word similarity and word analogy).}
\label{fig1}
\end{figure}

\subsection{RPD and Performance}

As \citet{Yin:18} discussed, usability of word embeddings, such as using them to solve analogy and relatedness tasks, is important to practitioners. 
Through applying different sets of word embeddings to word similarity and word analogy tasks \cite{Mikolov:13}, we study the relationship between RPD and word embeddings' performance. Specifically, we set the word embeddings produced by SGNS with 25 NS as a starting point and use other word embeddings, for example, GloVe as an end point. Then we get a two dimensional point with $x$ as their RPD, $y$ as their absolute performance change in word similarity\footnote{Our word similarity task can be found here: \url{ https://aclweb.org/aclwiki/WordSimilarity-353_Test_Collection_(State_of_the_art)}} and analogy\footnote{Our word analogy task can be found here: \url{ https://aclweb.org/aclwiki/Google_analogy_test_set_(State_of_the_art)}} tasks. 

By putting those points in Figure \ref{fig1}, we can tell in a certain range of RPD, the larger RPD between the two sets of word embedding means the bigger gap in their absolute performance. Intuitively, RPD is strongly related to cosine similarity, which is the measure of word similarity. RPD also shares the same property of PIP loss, where a small RPD leads to a small difference in relatedness and analogy tasks. We obtain similar results when the starting point is a different embedding space.

Note that this section serves to demonstrate the performance (at least in word similarity and analogy tasks) variation of different embedding spaces is correlated with their RPD. While we are aware of the relevance of other downstream tasks, we do not explore further since our focus lies in investigating the intrinsic geometry relation of embedding spaces.

\section{Experiment}
The following experiments serve to apply RPD to explore some questions of interest and further demonstrate that RPD is suitable for investigating the relations between embedding spaces. We leave applying RPD to help improve specific NLP tasks to future research. For example, RPD could be used for combining different embeddings together, which could help us produce better meta-embeddings \citep{kiela-etal-2018-dynamic}.
\subsection{Setup}

If not explicitly stated, the experiments are performed on Text8 corpus~\cite{matt:11}, a standard benchmark corpus used for various natural language tasks \cite{Yin:18}. For all methods we experiment, we train 300 dimension embeddings, with window size of 10, and normalize the embedding matrices with their standard deviation\footnote{The code can be found on Bitbucket: \url{https://bitbucket.org/omerlevy/hyperwords}}. The default NS for SGNS is 15. 

\subsection{Different Algorithms Produce Different Embeddings}

\subsubsection*{Dependence of SGNS and SVD$_\text{PMI}$}
As discussed in the introduction, the relationship between embeddings trained with SGNS and SVD$_\text{PMI}$ remains controversial~\citep{arora:16,mimno:17}. We use the results we obtain in Section 3.2 to test their dependence. For example, if one believes that $E_1$ trained with SGNS and $E_2$ trained with SVD$_\text{PMI}$ have no relationship, then the null hypothesis $H_0$ would be: $E_1$ and $E_2$ are independent.

Under $H_0$, $\text{RPD}(E_1,E_2)$ asymptotically follows $\mathcal{N}(\mu, \sigma^2)$. Then the test statistic $z$ is calculated as follows.
\[z = \frac{\text{RPD}(E_1, E_2)- \mu}{\sigma}\]

\begin{table}[t!]
\begin{center}
\ra{1.15}
\begin{tabular}{llll}
\toprule
 \bf Methods & \ GloVe & \ SVD$_\text{PMI}$ & \ SVD$_\text{LC}$\\ \midrule
SGNS$_{25}$ & 0.792 & 0.609 & 0.847\\
SGNS$_{15}$ & 0.773 & 0.594 & 0.837\\
SGNS$_5$& 0.725 & 0.550 & 0.805\\
SGNS$_1$ & 0.719 & 0.511 & 0.799\\
\bottomrule
\end{tabular}
\end{center}
\caption{RPDs of SGNS vs other methods}
\label{table1}
\end{table}

In our case, we estimate $\mu=0.953$ and $\sigma = 0.001$ with Monte Carlo simulation with randomly initialized embeddings. Take
RPD$(E_\text{SGNS$_{1}$}$, $E_\text{SVD$_\text{PMI}$})$ = 0.511
from Table \ref{table1} as an example, the statistic $z = 442$, which means the p-value $\ll$ 0.01. Thus, we can confidently reject $H_0$. Notice that we can test any two sets of word embeddings with this method. It is not hard to see that no pair of word embeddings in Table \ref{table1} are independent, which suggests that there exists an unified theory behind these methods.

\subsubsection*{SGNS is Closest to SVD$_\text{PMI}$}
With the help of RPD, it is also interesting to investigate distances between embeddings produced by different methods. Here, we calculate the RPDs among SGNS (with negative sampling 25, 15, 5, 1), GloVe, SVD$_\text{PMI}$, SVD$_\text{LC}$.

Table \ref{table1} shows the RPDs between SGNS with different negative sampling numbers and other methods. From the table, we can tell that SGNS stays close to SVD$_\text{PMI}$, which confirms \citet{Levy:14}'s theory. 

\begin{figure}[h]
\centering
\includegraphics[width=0.45\textwidth]{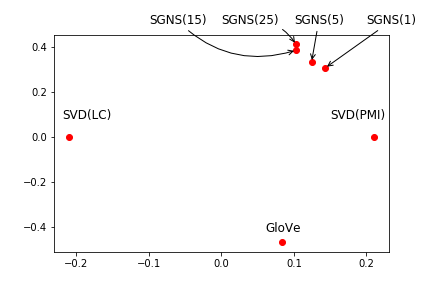}
\caption{Plot of different methods. We create the plot by fixing the position of SVD$_\text{LC}$ and SVD$_\text{PMI}$. We then derive the position of other word embeddings according to their RPD with existing points on the plot.}
\label{fig2}
\end{figure}

\subsubsection*{Hyper-parameters Have Influence on Embeddings}

From Table \ref{table1}, an interesting phenomenon is that SGNS becomes closer to other methods with the decrease of negative samples, which suggests that negative sampling is one of the factors driving SGNS away from matrix factorization methods. 

With RPDs between different sets of word embeddings, we plot the embeddings 
in 2D by treating each embedding space as a single point. We first fix point SVD$_\text{PMI}$ and SVD$_\text{LC}$, then we draw other points according to their RPDs with the other methods. Figure \ref{fig2} helps us see how negative sampling affects the embedding intuitively. Increasing the number of negative samples pulls SGNS away from SVD$_\text{PMI}$. Combining Table \ref{table1} and Figure \ref{fig2}, we can tell that although the hyper-parameters can influence the embeddings to some extent, the main difference comes from the algorithms. 

\subsection{Different Initializations Barely Influence Embeddings}

Random initializations produce different embeddings with the same algorithms and hyperparameters. While those embeddings usually get similar performance on the downstream tasks, people are still concerned about their effects. We investigate the influence of random initializations for GloVe and SGNS.

We train the embedding in the same setting multiple times and get the average RPDs for each method. For SGNS, the average RPDs of random initialization is 0.027. For GloVe, the value is 0.059.

We can tell that different random initializations produce essentially the same embeddings. Neither SGNS or GloVe has a significant RPD in different initializations, which suggests random initialization has little influence over word embeddings' performance (Section 3.4). However, SGNS seems to be more stable in this setting.

\subsection{Different Corpora Produce Different Embeddings}

It is well known that different 
corpora produce different word embeddings. However, it is hard for us to tell how different they are and whether the difference influences downstream applications~\cite{maria:18}. Knowing this would help researchers choose the algorithms in specific scenarios, for example, evolving semantic discovery \cite{10.1145/3159652.3159703,doi:10.1177/0003122419877135}. They focus on the semantic evolution of words, but corpora are different in
different time scales. Their methods use word embeddings to study semantic shift, which might be influenced by the word embeddings being trained on different corpora, thus getting unreliable results. In this case, it would be important to chose an algorithm less prone to influences by differences in
corpora.

We train word embeddings using each of text8~(Wikipedia domain, 25097 unique words), WMT14 news crawl\footnote{\url{http://www.statmt.org/wmt14/}}~(Newswire domain, 24359 unique words), TED speech\footnote{\url{https://workshop2016.iwslt.org/}}~(Speech domain, 7389 unique words). We compute RPD on the intersections of their 
vocabulary

\begin{table}[t!]
\begin{center}
\ra{1.15}
\begin{tabular}{lll}
\toprule   & \ SGNS & \ GloVe \\ \midrule
Text8-WMT14 & 0.168 & 0.686\\
Text8-TED & 0.119 & 0.758 \\
WMT14-TED & 0.175 & 0.716\\
\bottomrule
\end{tabular}
\end{center}
\caption{\label{corpus} RPDs between same method trained from different corpora }
\end{table}

From Table \ref{corpus}, we can tell that SGNS is consistently more stable than GloVe in different domains. We suggest that this is because GloVe trains the embedding with co-occurrence matrix, which gets influenced more by the corpus.

\section{Discussion}
While our work investigates some interesting problems about word embeddings, there are many other problems about embeddings that can be demonstrated with the help of RPD. We discuss some of them as follows.

\subsection{RPD and Crosslingual Word Embeddings} \citet{Artetxe18} provide a framework to obtain bilingual embeddings, whose the core step of the framework is an orthogonal transformation and other existing methods can be seen as its variations. The framework proposes to train monolingual embeddings separately and then map them into a shared-embedding space with linear transformation. 

While linear transformation is no guarantee for the alignment of two embedding spaces from different languages, RPD could potentially serve as a way to indicate how different language pairs benefit from mapping them with an orthogonal transformation. Since RPD is unitary-invariant, we can calculate RPD between embedding spaces from different language pairs. The smaller RPD is, the better the framework could align this two language embedding spaces. 

\subsection{RPD and Post-Processing Word Embeddings}
Post-processing word embeddings can be useful in many ways. For example, \citet{vulic18post} retrofit word embeddings with external linguistic resources, such as WordNet to obtain better embeddings; \citet{schutze16word} decompose embedding space to get better performance at specialized domains; and \citet{mu2018allbutthetop} obtain stronger embeddings by eliminating the common mean vector and a few top dominating directions.

RPD could serve as a metric to evaluate how the embedding space changes intrinsically after post-processing. 

\subsection{RPD and Contextualized Word Embeddings}

Contextualized embeddings are popular NLP techniques which significantly improve a wide range of NLP tasks \citep{bowman-etal-2015-large, rajpurkar-etal-2018-know}. To understand why contextualized embeddings are beneficial to those NLP tasks, many works investigate the the nature of syntactic \citep{liu-etal-2019-linguistic}, semantic \citep{liu-etal-2019-linguistic}, and commonsense knowledge \citep{zhou2019evaluating} contained in such representations. 

However, we still know little about the vector space of contextualized embeddings and their relationship with traditional word embeddings, which is important to further apply contextualized embeddings in various scenarios \citep{Lin2019SituatingSE}. RPD can potentially serve to help us better understand contextualized embeddings in future research.

\section{Conclusion}
In this paper, we propose RPD, a metric to quantify the distance between embedding 
spaces (i.e different sets of word embeddings). With the help of RPD and its properties, we verify some intuitions and answer some questions. Justifying RPD theoretically and empirically, we believe RPD can offer us a new perspective to understand and compare word embeddings.

\section*{Acknowledgments}
I would like to thank Dr. Zi Yin, Dr. Vered Shwartz, Maarten Sap, and Jorge Balazs for their feedback that greatly improved the paper.

\bibliography{acl2020}
\bibliographystyle{acl_natbib}

\appendix

\include{appendix}

\end{document}

%% file: appendix.tex
\theoremstyle{remark}

\newcommand\iid{i.i.d.}
\newcommand\pN{\mathcal{N}}

\section{Appendix A. Limitation of $||EE^T||$}
\label{appendix:a}
As discuss before, in our case, we can assume 
\iid~$v_{ij} \sim \pN(0, 1)$, where $v_{ij}$ is the $j^{th}$ entry in the $i^{th}$ word vector $v_i$ of $E$.

\begin{equation} \label{eq1}
\begin{split}
||EE^T|| & =n\sqrt{\frac{\sum_{i,j}^n(v_iv_j^T)^2}{n^2}}  \\
 & = n\sqrt{\frac{\sum_{i\neq j}^n(v_iv_j^T)^2}{n^2}+\frac{\sum_{i= j}^n(v_iv_j^T)^2}{n^2}}
\end{split}
\end{equation}

By the assumption, we know that $v_iv_j^T$ identically distributes for any $i\neq j, 1 \geq i \leq n, 1 \geq j \leq n $. By applying the law of large numbers, the term $\frac{\sum_{i\neq j}^n(v_iv_j^T)^2}{n^2}$ goes to $\mathbb{E}((v_iv_j^T)^2)$ as $n$ goes to $\infty$. The term $\frac{\sum_{i= j}^n(v_iv_j^T)^2}{n^2}$ goes to zero as $n$ goes to $\infty$. Then, we know that 
$||EE^T||\to n\sqrt{\mathbb{E}((v_iv_j^T)^2)}, n \to \infty$.

We only need to calculate $\mathbb{E}((v_iv_j^T)^2)$.
\begin{equation}
    \mathbb{E}((v_iv_j^T)^2) = Var(v_iv_j^T) + (\mathbb{E}(v_iv_j^T))^2
\end{equation}

Simple calculation shows that $Var(v_iv_j^T) = d$, $\mathbb{E}(v_iv_j^T) = 0$. Then $\mathbb{E}((v_iv_j^T)^2) =d $, $d$ is the dimension of word embedding here. Thus, $||EE^T|| \to n\sqrt{d}, n\to \infty$.

\section{Appendix B. Normality of RPD}
Let's review the form of RPD.
\begin{equation} \label{eq2}
\begin{split}
\text{RPD}(E_1,E_2)=\frac{1}{2} \frac{||E_1E_1^T-E_2E_2^T||^2}{||E_1E_1^T||||E_2E_2^T||}\\ 
\end{split}
\end{equation}

As we discuss in \ref{appendix:a},  $\frac{||E_1E_1^T||||E_2E_2^T||}{n^2} \to d$, as $n \to \infty$. We only have to prove $\frac{||E_1E_1^T-E_2E_2^T||^2}{n^2}$ distributes normally. The key is how to apply the central limit theorem~(CLT).

We denote as follows.
\begin{equation}
\begin{split}
    H_n & = \frac{||E_1E_1^T-E_2E_2^T||^2}{n^2}\\
    & =\frac{||E_1E_1^T||^2+||E_2E_2^T||^2-2\langle\, E_1E_1^T,E_2E_2^T\rangle}{n^2}
\end{split}
\end{equation}

Notice that the term $\langle\, E_1E_1^T,E_2E_2^T\rangle$ does not contribute to the variance if we analyze the second moment of the numerator. So it is equivalent to prove $T_n = \frac{||E_1E_1^T||^2+||E_2E_2^T||^2}{n^2} $ distributes normally. 

We project the $T_n$ to 

$S_n = \sum_{i,j}^n \mathbb{E}(T_n|v_{ij})-(n-1)\mathbb{E}(T_n)$

Simple calculation would show that $\frac{Var(T_n)}{Var(S_n)} \to 1, n \to \infty, \frac{n}{d} = c$.
Then by the Hajek projection theorem, we get $T_n$ has the same distribution as $S_n$. It is not hard to see that each random variable $\mathbb{E}(T_n|v_{ij})$ in $S_n$ is independent of others. This allows us to apply CLT to $S_n$ and get $S_n \sim \pN(\mu, \sigma^2)$. Thus, $H_n \sim \pN(\mu, \sigma^2)$. 

\section{Appendix C. Monte Carlo Simulation}
Here is how we perform Monte Carlo simulation. We independently produce two matrix $E_1, E_2 \in \mathbf{R}^{n \times d}$ with each entry i.i.d as $\pN(0,1)$. Then we calculate RPD($E_1, E_2$) and get the first RPD value. Repeat the process for 5000 times, we get a vector of RPDs. Drawing the histogram of this vector yields a normal distribution and we can estimate the mean and variance of the distribution by calculating the mean and variance of the vector of RPDs.

\section{Appendix D. Geometry Interpretation of RPD}
Now we consider a general case, where 
$\hat{E_1} $ and $\hat{E_2} $ are embeddings with n words.
\[
\begin{bmatrix} 
v_1^{(1)} \\ v_2^{(1)}\\ \vdots \\ v_n^{(1)}
\end{bmatrix}
,
\begin{bmatrix} 
v_1^{(2)} \\ v_2^{(2)}\\ \vdots \\ v_n^{(2)}
\end{bmatrix}
\]

Then

\begin{equation}
\begin{split}
 \frac{\langle\,\hat{E_1},\hat{E_2}\rangle}{||\hat{E_1}||||\hat{E_2}||}
 &=\frac{\sum_{i=1}^{n} v_i^{(1)T}v_i^{(2)}}
 {||\hat{E_1}||||\hat{E_2}||}\\
 &= \sum_{i=1}^{n} \frac{v_i^{(1)T}v_i^{(2)}}{||v_i^{1}||||v_i^{(2)}||}  \frac{||v_i^{1}||||v_i^{(2)}||}{||\hat{E_1}||||\hat{E_2}||}
\end{split}
\end{equation}

We denote $\frac{||v_i^{1}||||v_i^{(2)}||}{||\hat{E_1}||||\hat{E_2}||}$ as $w_i$, $\frac{v_i^{(1)T}v_i^{(2)}}{||v_i^{1}||||v_i^{(2)}||}$ as $\cos(\theta_{i})$

It is not hard to see that the $w_i \approx \frac{1}{n}$, when n is large enough. Then we get $\text{RPD}(E_1, E_2) \approx 1- \frac{\sum_{i=1}^n \cos(\theta_i)}{n}$. Considering the isotropic assumption again, another observation is that the $cos(\theta_i)$ distributes normally.

%% file: acl2020.bbl
\begin{thebibliography}{29}
\expandafter\ifx\csname natexlab\endcsname\relax\def\natexlab#1{#1}\fi

\bibitem[{Antoniak and Mimno(2018)}]{maria:18}
Maria Antoniak and David Mimno. 2018.
\newblock \href {https://transacl.org/ojs/index.php/tacl/article/view/1202}
  {Evaluating the stability of embedding-based word similarities}.
\newblock \emph{Transactions of the Association for Computational Linguistics},
  6:107--119.

\bibitem[{Arora et~al.(2016)Arora, Li, Liang, Ma, and Risteski}]{arora:16}
Sanjeev Arora, Yuanzhi Li, Yingyu Liang, Tengyu Ma, and Andrej Risteski. 2016.
\newblock \href {https://transacl.org/ojs/index.php/tacl/article/view/742} {A
  latent variable model approach to pmi-based word embeddings}.
\newblock \emph{Transactions of the Association for Computational Linguistics},
  4:385--399.

\bibitem[{Artetxe et~al.(2018)Artetxe, Labaka, and Agirre}]{Artetxe18}
Mikel Artetxe, Gorka Labaka, and Eneko Agirre. 2018.
\newblock \href
  {https://www.aaai.org/ocs/index.php/AAAI/AAAI18/paper/view/16935/16781}
  {Generalizing and improving bilingual word embedding mappings with a
  multi-step framework of linear transformations}.

\bibitem[{Bojanowski et~al.(2017)Bojanowski, Grave, Joulin, and
  Mikolov}]{bojanowski-etal-2017-enriching}
Piotr Bojanowski, Edouard Grave, Armand Joulin, and Tomas Mikolov. 2017.
\newblock \href {https://doi.org/10.1162/tacl_a_00051} {Enriching word vectors
  with subword information}.
\newblock \emph{Transactions of the Association for Computational Linguistics},
  5:135--146.

\bibitem[{Bolukbasi et~al.(2016)Bolukbasi, Chang, Zou, Saligrama, and
  Kalai}]{NIPS2016_6228}
Tolga Bolukbasi, Kai-Wei Chang, James~Y Zou, Venkatesh Saligrama, and Adam~T
  Kalai. 2016.
\newblock \href
  {http://papers.nips.cc/paper/6228-man-is-to-computer-programmer-as-woman-is-to-homemaker-debiasing-word-embeddings.pdf}
  {Man is to computer programmer as woman is to homemaker? debiasing word
  embeddings}.
\newblock In D.~D. Lee, M.~Sugiyama, U.~V. Luxburg, I.~Guyon, and R.~Garnett,
  editors, \emph{Advances in Neural Information Processing Systems 29}, pages
  4349--4357. Curran Associates, Inc.

\bibitem[{Bowman et~al.(2015)Bowman, Angeli, Potts, and
  Manning}]{bowman-etal-2015-large}
Samuel~R. Bowman, Gabor Angeli, Christopher Potts, and Christopher~D. Manning.
  2015.
\newblock \href {https://doi.org/10.18653/v1/D15-1075} {A large annotated
  corpus for learning natural language inference}.
\newblock In \emph{Proceedings of the 2015 Conference on Empirical Methods in
  Natural Language Processing}, pages 632--642, Lisbon, Portugal. Association
  for Computational Linguistics.

\bibitem[{Devlin et~al.(2019)Devlin, Chang, Lee, and
  Toutanova}]{devlin-etal-2019-bert}
Jacob Devlin, Ming-Wei Chang, Kenton Lee, and Kristina Toutanova. 2019.
\newblock \href {https://doi.org/10.18653/v1/N19-1423} {{BERT}: Pre-training of
  deep bidirectional transformers for language understanding}.
\newblock In \emph{Proceedings of the 2019 Conference of the North {A}merican
  Chapter of the Association for Computational Linguistics: Human Language
  Technologies, Volume 1 (Long and Short Papers)}, pages 4171--4186,
  Minneapolis, Minnesota. Association for Computational Linguistics.

\bibitem[{Kiela et~al.(2018)Kiela, Wang, and Cho}]{kiela-etal-2018-dynamic}
Douwe Kiela, Changhan Wang, and Kyunghyun Cho. 2018.
\newblock \href {https://doi.org/10.18653/v1/D18-1176} {Dynamic meta-embeddings
  for improved sentence representations}.
\newblock In \emph{Proceedings of the 2018 Conference on Empirical Methods in
  Natural Language Processing}, pages 1466--1477, Brussels, Belgium.
  Association for Computational Linguistics.

\bibitem[{Kozlowski et~al.(2019)Kozlowski, Taddy, and
  Evans}]{doi:10.1177/0003122419877135}
Austin~C. Kozlowski, Matt Taddy, and James~A. Evans. 2019.
\newblock \href {https://doi.org/10.1177/0003122419877135} {The geometry of
  culture: Analyzing the meanings of class through word embeddings}.
\newblock \emph{American Sociological Review}, 84(5):905--949.

\bibitem[{Levy and Goldberg(2014{\natexlab{a}})}]{Levy2014DependencyBasedWE}
Omer Levy and Yoav Goldberg. 2014{\natexlab{a}}.
\newblock \href {https://doi.org/10.3115/v1/P14-2050} {Dependency-based word
  embeddings}.
\newblock In \emph{Proceedings of the 52nd Annual Meeting of the Association
  for Computational Linguistics (Volume 2: Short Papers)}, pages 302--308,
  Baltimore, Maryland. Association for Computational Linguistics.

\bibitem[{Levy and Goldberg(2014{\natexlab{b}})}]{Levy:14}
Omer Levy and Yoav Goldberg. 2014{\natexlab{b}}.
\newblock \href {http://dl.acm.org/citation.cfm?id=2969033.2969070} {Neural
  word embedding as implicit matrix factorization}.
\newblock In \emph{Proceedings of the 27th International Conference on Neural
  Information Processing Systems - Volume 2}, NIPS'14, pages 2177--2185,
  Cambridge, MA, USA. MIT Press.

\bibitem[{Levy et~al.(2015)Levy, Goldberg, and Dagan}]{levy:15}
Omer Levy, Yoav Goldberg, and Ido Dagan. 2015.
\newblock \href {https://doi.org/10.1162/tacl\_a\_00134} {Improving
  distributional similarity with lessons learned from word embeddings}.
\newblock \emph{Transactions of the Association for Computational Linguistics},
  3:211--225.

\bibitem[{Lin and Smith(2019)}]{Lin2019SituatingSE}
Lucy~H. Lin and Noah~A. Smith. 2019.
\newblock \href {https://arxiv.org/abs/1909.10724} {Situating sentence
  embedders with nearest neighbor overlap}.
\newblock \emph{ArXiv}, abs/1909.10724.

\bibitem[{Liu et~al.(2019)Liu, Gardner, Belinkov, Peters, and
  Smith}]{liu-etal-2019-linguistic}
Nelson~F. Liu, Matt Gardner, Yonatan Belinkov, Matthew~E. Peters, and Noah~A.
  Smith. 2019.
\newblock \href {https://doi.org/10.18653/v1/N19-1112} {Linguistic knowledge
  and transferability of contextual representations}.
\newblock In \emph{Proceedings of the 2019 Conference of the North {A}merican
  Chapter of the Association for Computational Linguistics: Human Language
  Technologies, Volume 1 (Long and Short Papers)}, pages 1073--1094,
  Minneapolis, Minnesota. Association for Computational Linguistics.

\bibitem[{Mahoney(2011)}]{matt:11}
Matt Mahoney. 2011.
\newblock \href {http://mattmahoney.net/dc/textdata.html} {Large text
  comparison benchmark, 2011}.

\bibitem[{Mikolov et~al.(2013)Mikolov, Chen, Corrado, and Dean}]{Mikolov:13}
Tomas Mikolov, Kai Chen, Greg~S. Corrado, and Jeffrey Dean. 2013.
\newblock \href {http://arxiv.org/abs/1301.3781} {Efficient estimation of word
  representations in vector space}.

\bibitem[{Mimno and Thompson(2017)}]{mimno:17}
David Mimno and Laure Thompson. 2017.
\newblock \href {https://doi.org/10.18653/v1/D17-1308} {The strange geometry of
  skip-gram with negative sampling}.
\newblock In \emph{Proceedings of the 2017 Conference on Empirical Methods in
  Natural Language Processing}, pages 2873--2878.

\bibitem[{Mnih and Kavukcuoglu(2013)}]{Mnih2013LearningWE}
Andriy Mnih and Koray Kavukcuoglu. 2013.
\newblock \href
  {http://papers.nips.cc/paper/5165-learning-word-embeddings-efficiently-with-noise-contrastive-estimation.pdf}
  {Learning word embeddings efficiently with noise-contrastive estimation}.
\newblock In C.~J.~C. Burges, L.~Bottou, M.~Welling, Z.~Ghahramani, and K.~Q.
  Weinberger, editors, \emph{Advances in Neural Information Processing Systems
  26}, pages 2265--2273. Curran Associates, Inc.

\bibitem[{Mu and Viswanath(2018)}]{mu2018allbutthetop}
Jiaqi Mu and Pramod Viswanath. 2018.
\newblock \href {https://openreview.net/forum?id=HkuGJ3kCb} {All-but-the-top:
  Simple and effective postprocessing for word representations}.
\newblock In \emph{International Conference on Learning Representations}.

\bibitem[{Pennington et~al.(2014)Pennington, Socher, and
  Manning}]{pennington:14}
Jeffrey Pennington, Richard Socher, and Christopher Manning. 2014.
\newblock \href {https://doi.org/10.3115/v1/D14-1162} {{G}love: Global vectors
  for word representation}.
\newblock In \emph{Proceedings of the 2014 Conference on Empirical Methods in
  Natural Language Processing ({EMNLP})}, pages 1532--1543, Doha, Qatar.
  Association for Computational Linguistics.

\bibitem[{Peters et~al.(2018)Peters, Neumann, Iyyer, Gardner, Clark, Lee, and
  Zettlemoyer}]{peters-etal-2018-deep}
Matthew Peters, Mark Neumann, Mohit Iyyer, Matt Gardner, Christopher Clark,
  Kenton Lee, and Luke Zettlemoyer. 2018.
\newblock \href {https://doi.org/10.18653/v1/N18-1202} {Deep contextualized
  word representations}.
\newblock In \emph{Proceedings of the 2018 Conference of the North {A}merican
  Chapter of the Association for Computational Linguistics: Human Language
  Technologies, Volume 1 (Long Papers)}, pages 2227--2237, New Orleans,
  Louisiana. Association for Computational Linguistics.

\bibitem[{Rajpurkar et~al.(2018)Rajpurkar, Jia, and
  Liang}]{rajpurkar-etal-2018-know}
Pranav Rajpurkar, Robin Jia, and Percy Liang. 2018.
\newblock \href {https://doi.org/10.18653/v1/P18-2124} {Know what you don{'}t
  know: Unanswerable questions for {SQ}u{AD}}.
\newblock In \emph{Proceedings of the 56th Annual Meeting of the Association
  for Computational Linguistics (Volume 2: Short Papers)}, pages 784--789,
  Melbourne, Australia. Association for Computational Linguistics.

\bibitem[{Rothe and Sch{\"u}tze(2016)}]{schutze16word}
Sascha Rothe and Hinrich Sch{\"u}tze. 2016.
\newblock \href {https://doi.org/10.18653/v1/P16-2083} {Word embedding calculus
  in meaningful ultradense subspaces}.
\newblock In \emph{Proceedings of the 54th Annual Meeting of the Association
  for Computational Linguistics (Volume 2: Short Papers)}, pages 512--517,
  Berlin, Germany. Association for Computational Linguistics.

\bibitem[{Smith et~al.(2017)Smith, Turban, Hamblin, and Hammerla}]{smith:17}
Samuel~L. Smith, David H.~P. Turban, Steven Hamblin, and Nils~Y. Hammerla.
  2017.
\newblock \href {http://arxiv.org/abs/1702.03859} {Offline bilingual word
  vectors, orthogonal transformations and the inverted softmax}.
\newblock \emph{CoRR}, abs/1702.03859.

\bibitem[{Vuli{\'c} et~al.(2018)Vuli{\'c}, Glava{\v{s}}, Mrk{\v{s}}i{\'c}, and
  Korhonen}]{vulic18post}
Ivan Vuli{\'c}, Goran Glava{\v{s}}, Nikola Mrk{\v{s}}i{\'c}, and Anna Korhonen.
  2018.
\newblock \href {https://doi.org/10.18653/v1/N18-1048} {Post-specialisation:
  Retrofitting vectors of words unseen in lexical resources}.
\newblock In \emph{Proceedings of the 2018 Conference of the North {A}merican
  Chapter of the Association for Computational Linguistics: Human Language
  Technologies, Volume 1 (Long Papers)}, pages 516--527, New Orleans,
  Louisiana. Association for Computational Linguistics.

\bibitem[{Yao et~al.(2018)Yao, Sun, Ding, Rao, and
  Xiong}]{10.1145/3159652.3159703}
Zijun Yao, Yifan Sun, Weicong Ding, Nikhil Rao, and Hui Xiong. 2018.
\newblock \href {https://doi.org/10.1145/3159652.3159703} {Dynamic word
  embeddings for evolving semantic discovery}.
\newblock In \emph{Proceedings of the Eleventh ACM International Conference on
  Web Search and Data Mining}, WSDM ’18, page 673–681, New York, NY, USA.
  Association for Computing Machinery.

\bibitem[{Yin and Sch{\"u}tze(2016)}]{yin-schutze-2016-learning}
Wenpeng Yin and Hinrich Sch{\"u}tze. 2016.
\newblock \href {https://doi.org/10.18653/v1/P16-1128} {Learning word
  meta-embeddings}.
\newblock In \emph{Proceedings of the 54th Annual Meeting of the Association
  for Computational Linguistics (Volume 1: Long Papers)}, pages 1351--1360,
  Berlin, Germany. Association for Computational Linguistics.

\bibitem[{Yin and Shen(2018)}]{Yin:18}
Zi~Yin and Yuanyuan Shen. 2018.
\newblock \href
  {http://papers.nips.cc/paper/7368-on-the-dimensionality-of-word-embedding.pdf}
  {On the dimensionality of word embedding}.
\newblock In S.~Bengio, H.~Wallach, H.~Larochelle, K.~Grauman, N.~Cesa-Bianchi,
  and R.~Garnett, editors, \emph{Advances in Neural Information Processing
  Systems 31}, pages 887--898. Curran Associates, Inc.

\bibitem[{Zhou et~al.(2019)Zhou, Zhang, Cui, and Huang}]{zhou2019evaluating}
Xuhui Zhou, Yue Zhang, Leyang Cui, and Dandan Huang. 2019.
\newblock \href {http://arxiv.org/abs/1911.11931} {Evaluating commonsense in
  pre-trained language models}.

\end{thebibliography}
